\documentclass[conference]{IEEEtran}
\IEEEoverridecommandlockouts
\usepackage{cite}
\usepackage{amsmath,amssymb,amsfonts}
\usepackage{algorithmic}
\usepackage{graphicx}
\usepackage{textcomp}
\usepackage{xcolor}
\usepackage{subfigure}
\usepackage{pgfplots}
\usepackage{multirow}
\usepackage{booktabs}
\pgfplotsset{compat=1.17}

\def\BibTeX{{\rm B\kern-.05em{\sc i\kern-.025em b}\kern-.08em
    T\kern-.1667em\lower.7ex\hbox{E}\kern-.125emX}}
\begin{document}

\title{AugDMC: Data Augmentation Guided Deep Multiple Clustering
}


\author{\IEEEauthorblockN{Jiawei Yao, Enbei Liu, Maham Rashid, Juhua Hu}

\IEEEauthorblockA{\textit{School of Engineering and Technology, University of Washington, Tacoma, WA 98402, USA}}
}

\maketitle

\begin{abstract}

Clustering aims to group similar objects together while separating dissimilar ones apart. Thereafter, structures hidden in data can be identified to help users understand data in an unsupervised manner. Traditional clustering methods such as k-means provide only a single clustering for one data set. Deep clustering methods such as auto-encoder based clustering methods have shown a better performance, but still provide a single clustering. However, a given dataset might have multiple clustering structures and each represents a unique perspective of the data. 
Therefore, some multiple clustering methods have been developed to discover multiple independent structures hidden in data. Although deep multiple clustering methods provide better performance, how to efficiently capture a user's interest is still a problem. 
In this paper,
we propose AugDMC, a novel data Augmentation guided Deep Multiple Clustering method, to tackle the challenge. 
Specifically, AugDMC leverages data augmentations to automatically extract features related to a certain aspect of the data using a self-supervised prototype-based representation learning, where different aspects of the data can be preserved under different data augmentations. Moreover, a stable optimization strategy is proposed to alleviate the unstable problem from different augmentations. Thereafter, multiple clusterings based on different aspects of the data can be obtained.
Experimental results on two real-world datasets compared with state-of-the-art methods validate the effectiveness of the proposed method.

\end{abstract}


\section{Introduction}
Clustering, which clusters data points according to their similarities, has been intensively studied in the data mining and machine learning community.
Traditional clustering algorithms, such as k-means~\cite{macqueen1967some}, spectral clustering~\cite{ng2001spectral}, and Gaussian mixture model~\cite{bishop2006pattern}, group data into distinct collections with handcrafted features. However, these features are designed for general purposes and unsuitable for specific tasks. With the development of deep learning, deep clustering algorithms~\cite{xie2016unsupervised,peng2017cascade,ghasedi2017deep,guerin2018improving}
adopt Deep Neural Networks (DNNs) to perform clustering, showing dramatic improvement in clustering performance.
They have been widely applied in different applications such as 
bioinformatics~\cite{guzzi2014discussion}, 
computer vision~\cite{abd2021automatic}, 
and
text mining~\cite{lydia2018document,abualigah2020nature}.

\begin{figure}
    \centering
    \includegraphics[width=\linewidth]{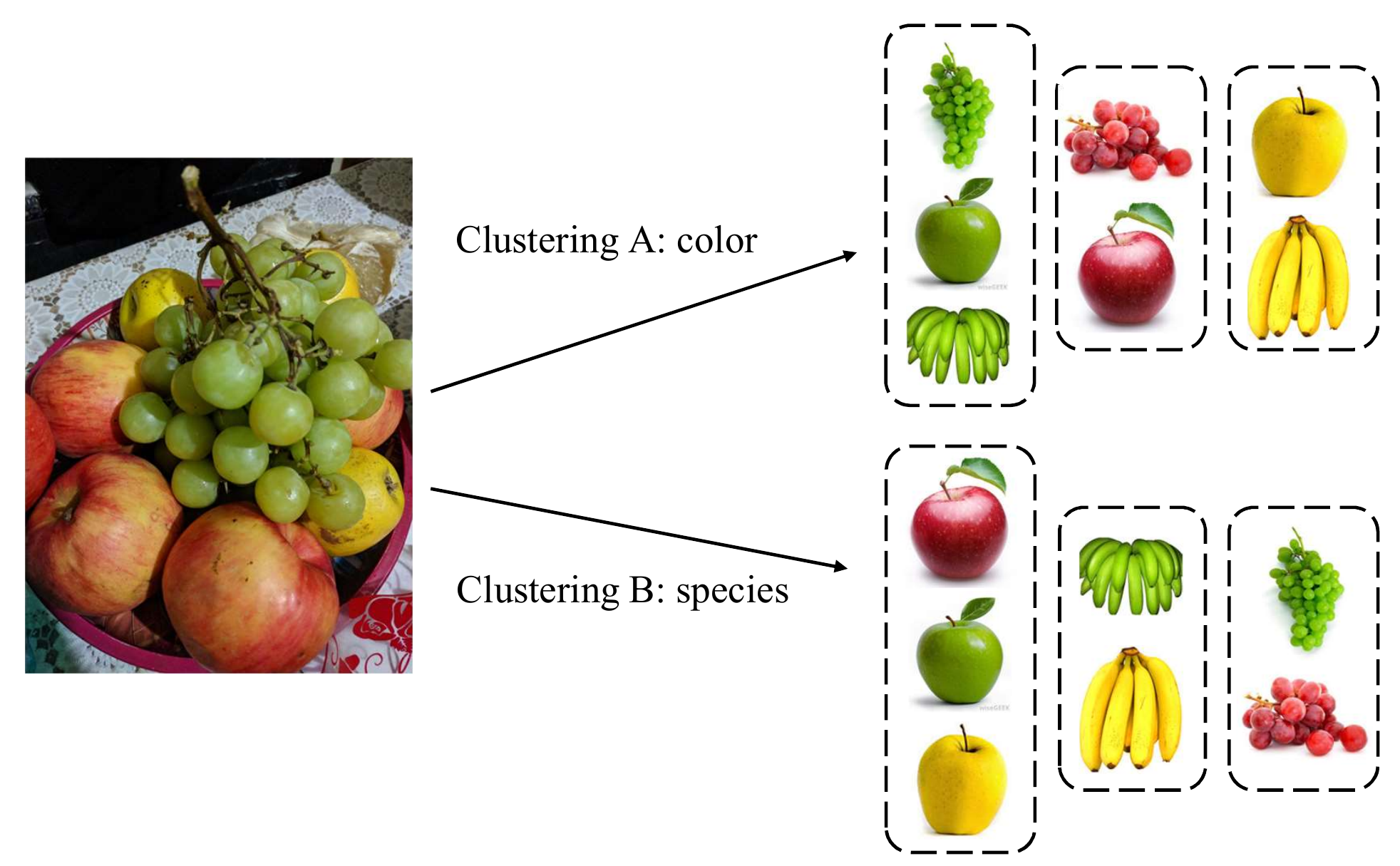}
    \caption{An example of multiple clusterings. Fruits can be clustered differently by color and species.}
    \label{fig:intro_fruit}
\end{figure}

Most clustering algorithms produce a single group of data partitions.
However, in real-world applications,
there might have different orthogonal ways to partition a given dataset. 
For example,
as shown in Fig.~\ref{fig:intro_fruit},
fruits have two orthogonal ways of data partitions: color and species.
To address this problem,
some researchers propose multiple clustering algorithms that aim to find more than one way to partition a given dataset.
For instance, 
MSC~\cite{hu2017finding} demonstrates the correlation between Laplacian eigengap and stability, and uncovers multiple clusterings by maximizing the eigengap.
MVMC~\cite{yao2019multi} leverages Hilbert-Schmidt Independence Criterion (HSIC) 
as a constraint to enhance the diversity between clusterings. Multiple clustering has also been found useful in real applications, such as epistatic interaction detection~\cite{wang2021epimc}.

Recently, some attempt to do deep multiple clustering to enhance the performance, e.g., ENRC~\cite{miklautz2020deep} combines auto-encoder and clustering objective function to obtain alternative clusterings, which has proven to give better clustering results than shallow multiple clustering methods. 
However,
capturing an end user’s interest, which corresponds to some specific perspectives of the data, is still a challenge in multiple clustering.
For example, if we ask ENRC~\cite{miklautz2020deep} to generate only two different clusterings, both of them may not be within an end user's interest. Therefore, we have to produce more. Thereafter, the end user has to manually check them one by one to choose those ones of interest, which makes the applicability in real-world tasks a problem. Therefore, in this paper, we aim to not only take the benefit of deep multiple clustering for better performance, but also generate only clusterings within users' interest.

Data augmentation~\cite{van2001art} has been widely used as a technique to enhance the generalization performance. However, its ability to capture different aspects of the data for multiple clustering has been ignored. For example, no matter how we change the color of an image, the shape of the interested object is always preserved. Based on this observation, we propose to use different data augmentation methods to preserve different aspects of the data, which can also be used to capture a user's interest to guide the generation of different clusterings. Concretely, we propose a novel deep multiple clustering method called data Augmentation guided Deep Multiple Clustering (AugDMC), in which 
we leverage prototype-based self-supervised learning to obtain different data representations guided by different augmentation methods. Thereafter, representations from each data perspective can be feed to any single clustering method to obtain a clustering.
We also propose a stable optimization strategy to ensure that the learned representations are robust when multiple data augmentations are applied. 
The main contributions of this paper are highlighted as follows:
\begin{itemize}
    \item We study a novel problem of multiple clustering to capture users' interest efficiently, which aims to control the aspect of clusterings via data augmentation. To the best of our knowledge, we are the first to study this problem.
    \item We propose AugDMC, a novel deep multiple clustering method guided by data augmentations. The proposed method uses a prototype-based self-supervised representation learning to obtain image representations for clustering and control the aspect of clusterings through data augmentations.
    \item The experimental results on three real-world datasets demonstrate the effectiveness of the proposed method, compared with the state-of-the-art methods. 
\end{itemize}

\section{Related Work}


\subsection{Multiple Clustering}
Traditional multiple clustering methods~\cite{hu2018subspace} leverage shallow models to discover alternative clustering in data.
Some of these methods use constraints to generate alternative clusterings.
For example, COALA~\cite{bae2006coala} 
treats objects within an established clustering as constraints for generating an alternative clustering.
Qi et al.~\cite{qi2009principled} regarded multiple clustering as a constrained optimization problem to obtain an alternative clustering.
Besides,
some methods are relying on different feature subspaces.
As an instance, Hu et al.~\cite{hu2017finding} proved the relation between Laplacian eigengap and stability of a clustering, and discovered multiple clusterings via maximizing the eigengap within different feature subspaces.
Inspired by non-negative matrix factorization~\cite{lee1999learning},
MNMF~\cite{yang2017non} incorporates the inner product of similarity matrices  as a regularization to generate multiple clusterings.
Moreover,
information theory is also used by some multiple clustering methods.
Gondek and Hofmann~\cite{gondek2003conditional} leveraged conditional information bottleneck to generate alternative clustering.
Dang and Bailey~\cite{dang2010generation} used an expectation maximization framework to optimize mutual information to obtain multiple clusterings.

Recently,
some researchers leveraged deep learning to generate multiple clusterings and achieved better results.
Wei et al.~\cite{wei2020multi} proposed a deep matrix factorization based method to discover multiple clusterings using multi-view data.
Besides,
ENRC~\cite{miklautz2020deep} uses an auto-encoder to learn the object features and generates multiple clusterings via optimizing a clustering objective function.
Since multi-head attention could learn features from different aspects~\cite{vaswani2017attention},
iMClusts~\cite{ren2022diversified} makes use of auto-encoders and multi-head attention to generate multiple clusterings. Although better performance has been achieved by these deep multiple clustering methods, they lack the flexibility and efficiency to capture a user's interest. That is, generating limited number of clusterings may not be able to cover a user's interest, but generating too many clusterings overloads a user's manual check, which is tackled in this work.

\subsection{Image Augmentation}
Data augmentation is greatly important to overcome the limitation of data samples~\cite{van2001art}. Image augmentation has achieved good results in downstream tasks like image classification~\cite{fawzi2016adaptive,mikolajczyk2018data},
image segmentation~\cite{nalepa2019data,zhao2019data},
object detection~\cite{jo2017data,zoph2020learning}.
Traditional image data augmentation consists of geometric transformation and photometric shifting.
Flipping~\cite{vyas2018fundamentals} maximizes the number of images in a dataset by reflecting an image around its vertical axis or horizontal axis.
Rotation~\cite{sifre2013rotation} generates images by rotating the image around an axis either in the right or left direction.
Cropping~\cite{sifre2013rotation} is a process of magnifying the original image via cutting and scaling images.
Color space shifting~\cite{winkler2005} is to shift color space, e.g., RGB, CMY, YIQ, HSL, to generate images.
Image filters~\cite{galdran2017data} are image processing techniques to augment images, e.g, histogram equalization, brightness increases, sharpening, blurring, and filters.

Besides,
some data augmentation methods are  based on deep learning.
Deep learning image data augmentation consists of three main categories, i.e., 
generative adversarial networks (GAN), neural style transfer (NST), and meta metric learning.
GAN-based methods generate artificial images from the initial dataset and utilize them to predict features of the images~\cite{yi2019generative}.
For the NST-based methods, they leverage neural representations of material and style to isolate and recombine pictures, demonstrating a way to construct creative images computationally~\cite{gatys2015neural}.
Meta metric learning methods leverage models with meta-learning architecture to generate images~\cite{zoph2016neural}.
A comprehensive review of image augmentation can be found in the surveys~\cite{shorten2019survey,khalifa2022comprehensive}.

Recently,
some researchers tried to combine clustering with augmentation.
Guo et al.~\cite{guo2018deep} proposed a clustering with data augmentation method named DEC-DA,
which designs an auto-encoder with augmented data and optimizes it via a clustering loss.
ASPC-DA~\cite{guo2019adaptive} trains an auto-encoder with augmented data and fine-tunes the model with a self-paced strategy.
Abavisani et al.~\cite{abavisani2020deep} designed some
efficient data augmentation policies and proposed a subspace clustering method using the augmented images.
Although image augmentation has shown its powerful ability in clustering,
these methods are not designed for multiple clustering. More importantly, the observation that different augmentation methods can help preserve different aspects of the data was ignored, which is studied in this paper.

\begin{figure*}
    \centering
    \includegraphics[width=\linewidth]{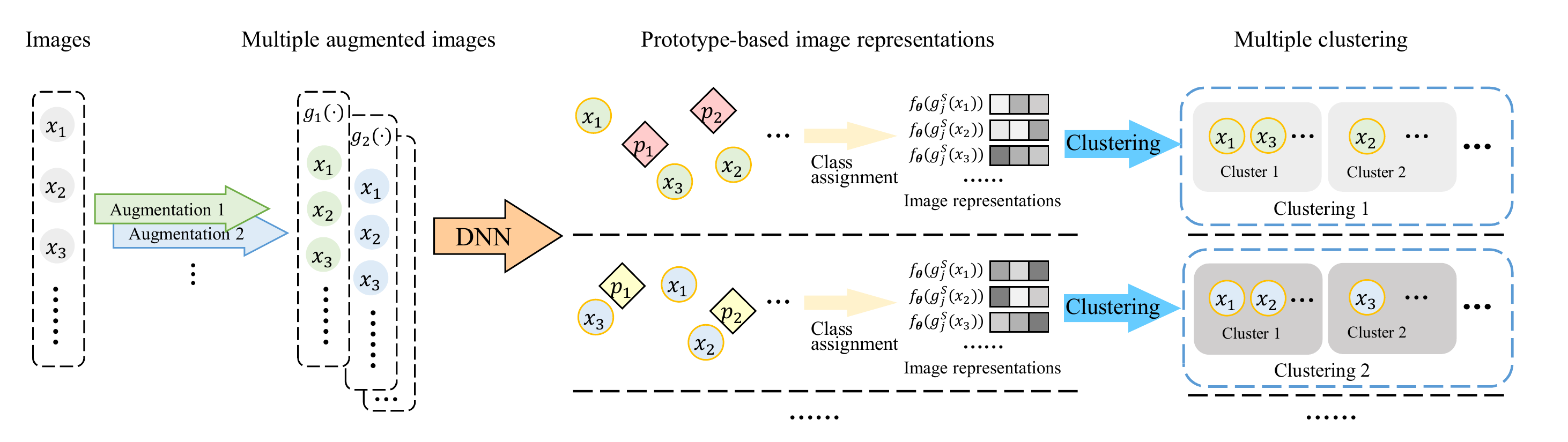}
    \caption{The framework of AugDMC. AugDMC uses multiple augmentation methods to obtain augmented images with desired characteristics. The representations of the augmented images are learned via a self-supervised prototype-based representation learning method. The final multiple clusterings can be obtained by employing any single clustering algorithm on the learned representations.}
    \label{fig:model}
\end{figure*}

\section{The Proposed Method}

To find multiple clustering structures hidden in the data with the flexibility and efficiency to capture a user's interest, we propose a novel data augmentation guided deep multiple clustering method, named AugDMC.

\subsection{Self-supervised prototype-based image representation}\label{ins}

The first step of AugDMC is to learn image representations,
which aims to learn a map $f_{\boldsymbol{\theta}}(x_i)$
without supervision, where
$f_{\boldsymbol{\theta}}(\cdot)$ is a deep neural network,
mapping image $x_i$ to a $d$-dimensional feature representation $f_{\boldsymbol{\theta}}(x_i)\in \mathbb{R}^{d}$ (all notations are summarized in Table~\ref{tab:notations}). 

In this paper, we propose a prototype-based latent class assignment strategy to learn the data representations. 
\begin{table}
    \centering
    \caption{Notations and Explanations.}
    \resizebox{\linewidth}{!}{
        \begin{tabular}{c | c}
        \toprule
        Notation    &   Explanation\\
         \midrule
         $x_i$      &   an image $i$\\
        $\tau$ & temperature parameter \\
        $n$ &   the total number of images in a dataset \\
        $K$ &   the total number of prototypes  \\
         $f_{\boldsymbol{\theta}}(\cdot)$   & a map via a deep neural network with parameters $\boldsymbol{\theta}$\\
        $f_{\boldsymbol{\theta}}(x_i)$ & mapping image $x_i$ to a feature representation \\
        $\boldsymbol{p}_k$ & a prototype, corresponding to latent class $k$\\
        $g_j(\cdot)$ & an augmentation method $j$ \\
        $g_j(x_i)$  & augmented image of $x_i$ by augmentation method $j$\\
        $\boldsymbol{y}$ & the set of ground truth labels in clustering \\
        $\boldsymbol{c}$ & the set of predicted labels in clustering \\
         \bottomrule
        \end{tabular}
    }
    \vspace{-0.8cm}
    \label{tab:notations}
\end{table}
Specifically,
let $\{\boldsymbol{p}_k\}_{k=1}^{K}$ indicate $K$ prototypes that describe $K$ anchors of the latent classes in the dataset.
$\boldsymbol{p}_k\in\mathbb{R}^d$ is a $d$-dimensional prototype, corresponding to latent class $k$ in the dataset. The similarity between image $x_i$ and prototype $k$
can be measured by the inner product between $f_{\boldsymbol{\theta}}(x_i)$ and $\boldsymbol{p}_k$ as
    $s_{ij} = \boldsymbol{p}_k^T \cdot f_{\boldsymbol{\theta}}(x_i)$.  
Thus, 
the probability of
image $x_i$ belonging to latent class $k$ can be described as
\begin{equation}\label{eq:probability}
    P(k|{x}_i;\boldsymbol{\theta}) = \frac{\exp{(\boldsymbol{p}_k^T \cdot f_{\boldsymbol{\theta}}(x_i)/\tau)}}{\sum_{k=1}^{K}\exp(\boldsymbol{p}_k^T \cdot f_{\boldsymbol{\theta}}(x_i)/\tau)},
\end{equation}
where $\tau$ is a temperature parameter that controls the scale of values,
so as to control the concentration level of the probability distribution~\cite{hinton2015distilling}.

Therefore,
considering all images in a dataset,
the objective function of the proposed method is to maximize the joint probability as
    $\prod_{i=1}^{n} P(k|{x}_i;{\boldsymbol{\theta}} )$,
where $n$ is the number of images in the dataset
and $\boldsymbol{\theta}$ indicates the parameters of the deep neural network $f_{\boldsymbol{\theta}}(\cdot)$. The deep neural network used in AugDMC consists of multiple convolutional layers and a single fully-connected (FC) layer.
Therefore, 
AugDMC is very flexible to employ these neural networks, such as
ResNet~\cite{he2016deep}, MobileNet~\cite{howard2017mobilenets}, EfficientNet~\cite{tan2019efficientnet}, etc.

\subsection{Augmentation}
AugDMC leverages augmentation to obtain images that could reflect different characteristics.
Given an image $x_i$, 
it can be augmented by a function $g_j(\cdot)$,
so the representation of augmented image $g_j(x_i)$
can be denoted as $f_{\boldsymbol{\theta}}\big(g_j(x_i)\big)$.
Considering the prototype-based representation learning, Eqn.~\eqref{eq:probability} can be rewritten as
\begin{equation}\label{eq:probability_aug}
    P(k|{x}_i;\boldsymbol{\theta}) = \frac{\exp{\big(\boldsymbol{p}_k^T \cdot f_{\boldsymbol{\theta}}\big(g_j(x_i)\big)/\tau\big)}}{\sum_{k=1}^{K}\exp\big(\boldsymbol{p}_k^T \cdot f_{\boldsymbol{\theta}}\big(g_j(x_i)\big)/\tau\big)}.
\end{equation}
Therefore,
the joint probability of all the images has the formulation
    $\prod_{i=1}^{n} P(k|g_j({x}_i);{\boldsymbol{\theta}} )$.
To sum up,
prototype-based representation learning aims to learn discriminative representations,
while augmentation provides the invariant property for a given perturbation. AugDMC could discover different aspects of representations from the combination.


\subsection{Multiple Clusterings from Multiple Augmentations}
Discrimination is essential for an effective representation learning, 
which can be captured by our prototype-based image representation learning. 
However, 
images can be separated in different ways. Therefore, we propose to do clustering that can aggregate similar ones according to the invariant property based on the augmentation. 
Specifically,
given a set of augmentations $\{g_1,\dots, g_J\}$, 
the invariant property corresponding to a certain augmentation is
\begin{equation}
    \min_f \|f(g_j({x}_i)) - f({x}_i)\|_2.
\end{equation}
With appropriate augmentations, we can learn multiple aspects of the data under different invariant properties. 
Thereafter, multiple clusterings can be realized by employing multiple augmentations.

However, one major challenge is to identify effective augmentations for multiple clusterings. 
Note that there are some prevalent properties for multiple clusterings, e.g., color, shape, etc~\cite{shorten2019survey,khalifa2022comprehensive}. 
To identify which augmentations should be employed,
one straightforward approach is to directly leverage standard augmentation for color invariant (e.g., color jitter), shape invariant (e.g., crop, rotation), etc.
This is however very inefficient, since it is hard to choose the color and angle that should be used. AugDMC aims to learn augmentation setups simultaneously.
Specifically,
for color-invariant property,
given a set of images, 
AugDMC extracts dominating colors from each image and then perturbs with extracted colors for a more effective color jitter augmentation. 
For shape-invariant property,
which can be obtained by crop and rotation.
AugDMC computes the pixel difference between original images from different angles of rotation and keeps those angles with the largest variance.





\subsection{Stable Optimization Strategy}
Although AugDMC can achieve multiple clusterings via multiple augmentations,
it can suffer from an unstable learning procedure caused by augmentation~\cite{cubuk2020randaugment}.
In other words, 
multiple augmentations applied to a single image may result in unstable representations without label information.
To address this problem,
inspired by RandAug~\cite{cubuk2020randaugment} that randomly selects a subset of augmentations at each iteration in the training process, we further design a stable optimization strategy for AugDMC.
Specifically,
we randomly draw a subset $\{g_j\}_{j=1}^S$ from $\{g_j\}_{j=1}^J$,  
and then learn the prototype-based representations with the selected augmentations.
Thus,
the final joint probability can be written as
\begin{equation}\label{eq:loss_prod_s}
    \prod_{i=1}^{n} P(k|g_j^S({x}_i);{\boldsymbol{\theta}} )=\prod_{i=1}^{n} P(k|g_1(g_2(\dots(g_S({x}_i))));{\boldsymbol{\theta}}).
\end{equation}
Since
a stable clustering can be obtained when images can be separated well,
we set the number of latent classes $K$ to be the number of images in the prototype-based 
representation learning process to maximize representation discrimination. 
In addition,
we repeat the process with a fixed number of epochs and only keep the feature extractors that achieve a satisfied performance,
i.e., 
the accuracy or loss of the prototype-based learning is not changed in a fixed number of epochs.
Then, multiple clusterings will be obtained using different feature extractors.
It is worth noting that although AugDMC adopts a random strategy to select augmentation methods, it can still use different augmentation candidate sets to make AugDMC capture specific features. For example, using rotation, flip, or crop can capture color-related features, while using colorjitter or grey can capture shape-related features.

Note that
the objective of AugDMC in Eqn.~\eqref{eq:loss_prod_s} is equivalent to minimizing the negative log-likelihood, 
so the final objective function of the proposed method is
\begin{equation}
    l(\boldsymbol{\theta})=-\sum_{i=1}^n\log P(k|g_j^S({x}_i);{\boldsymbol{\theta}} ).
\end{equation}
Thereafter, we can feed learned representations from each feature extractor to a clustering algorithm,  
such as k-means~\cite{lloyd1982least}, to obtain a clustering result, where multiple feature extractors provide multiple clusterings. Fig.~\ref{fig:model} summarizes the procedure of our proposal.

\begin{figure*} [ht]
    \centering
    \subfigure[MSC color]{
        \includegraphics[width=0.22\textwidth]{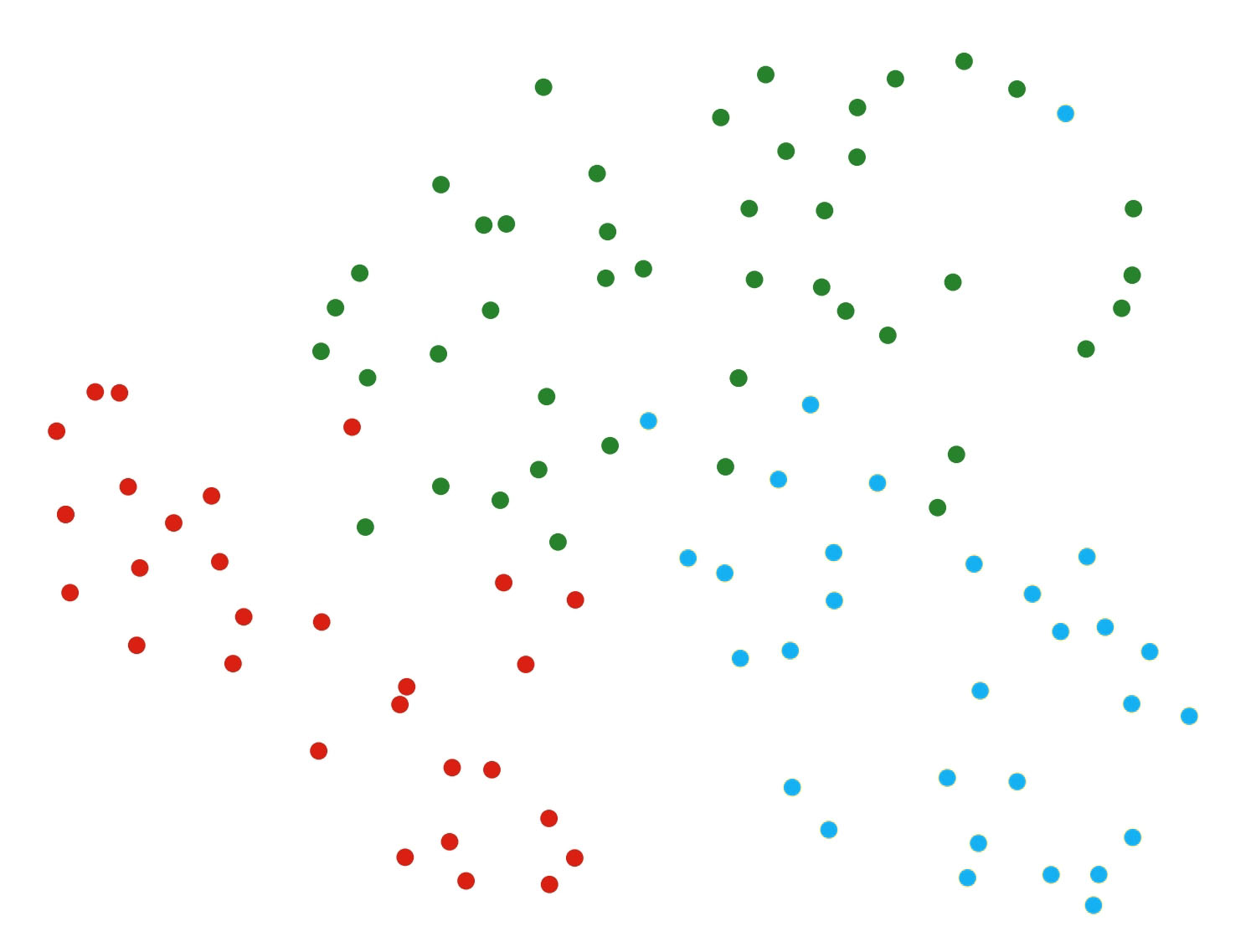}
    }
    \subfigure[MCV color]{
        \includegraphics[width=0.22\textwidth]{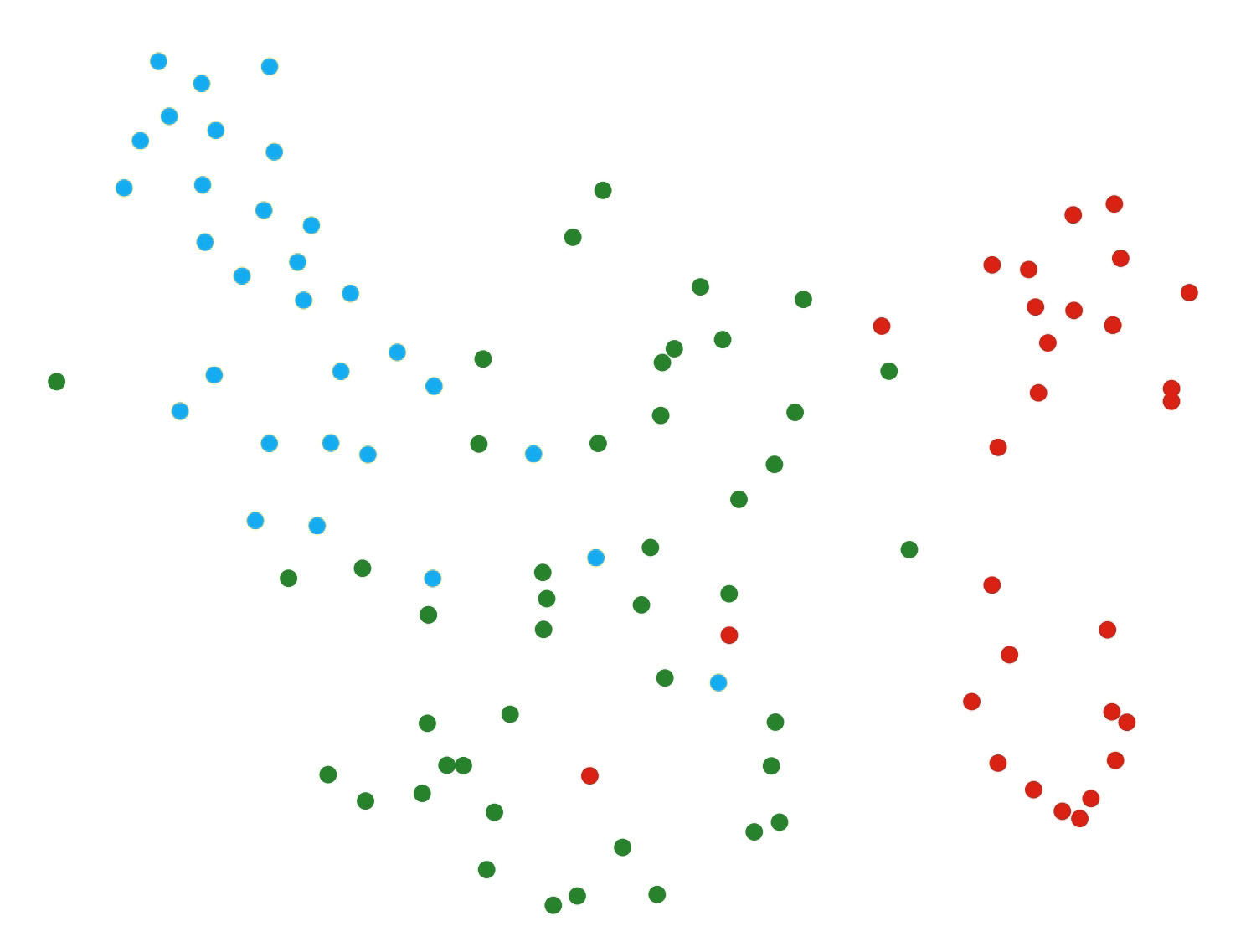}
    }
    \subfigure[ENRC color]{
        \includegraphics[width=0.22\textwidth]{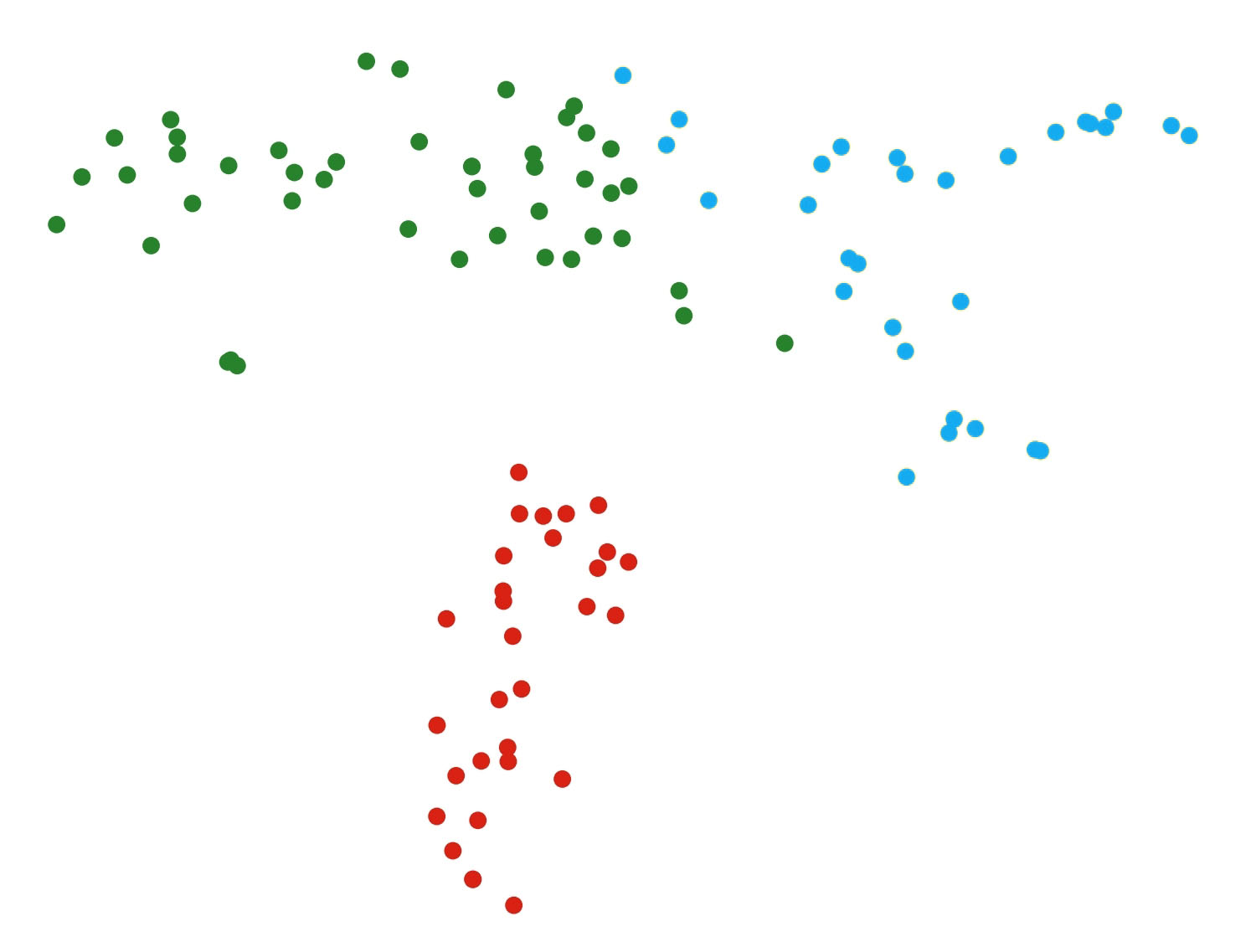}
    }
    \subfigure[AugDMC color]{
        \includegraphics[width=0.22\textwidth]{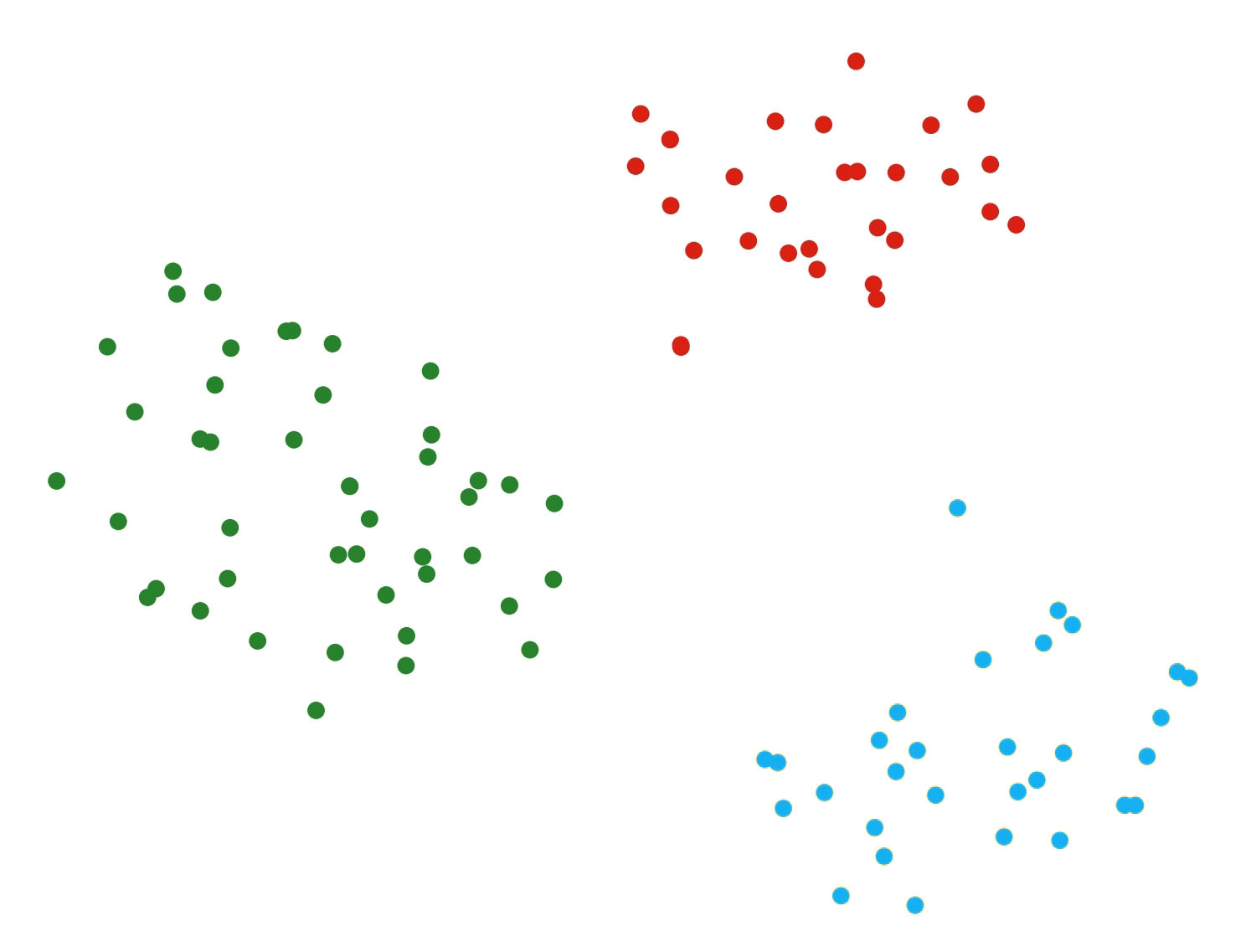}
    }\\
    \subfigure[MSC species]{
        \includegraphics[width=0.22\textwidth]{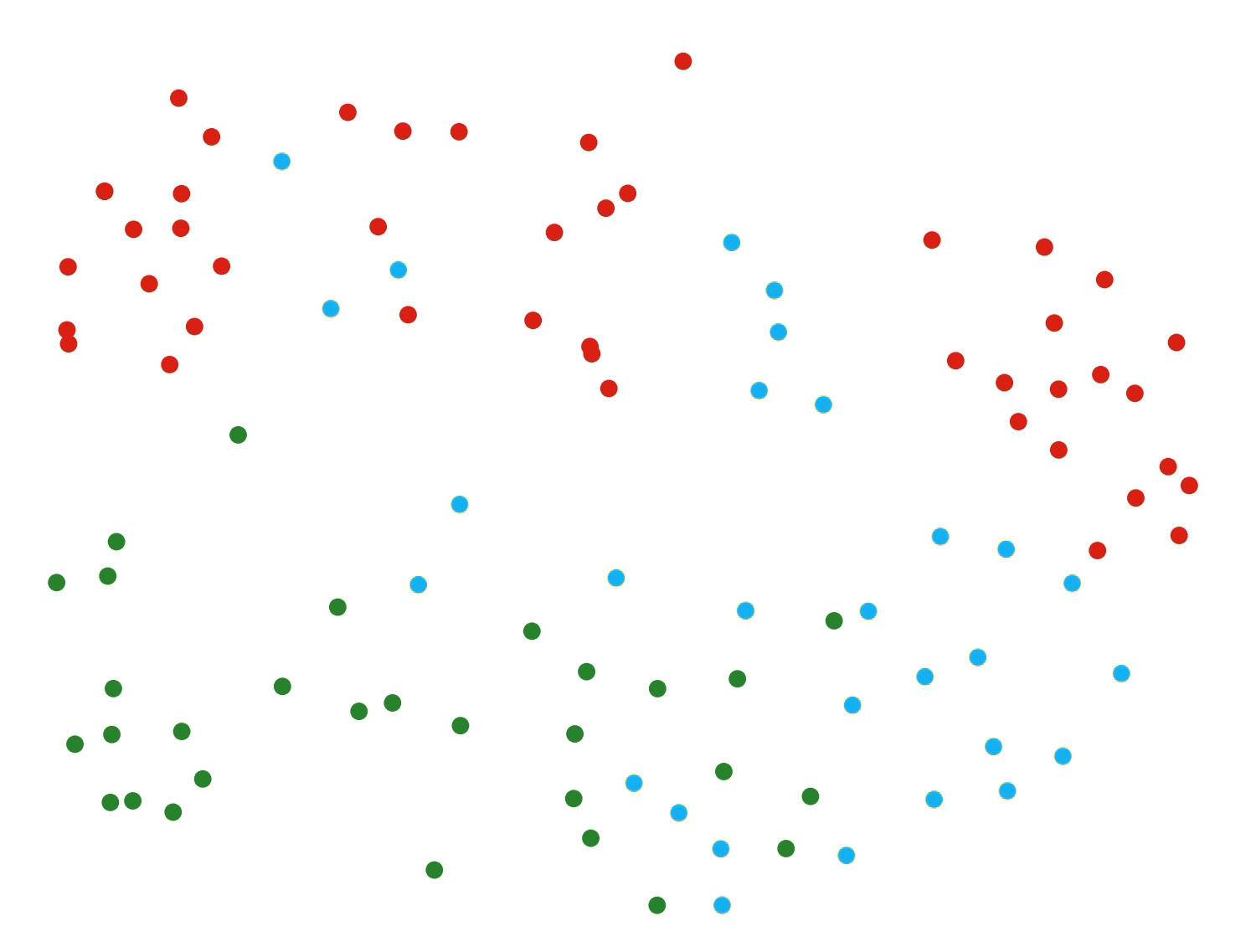}
    }
    \subfigure[MCV species]{
        \includegraphics[width=0.22\textwidth]{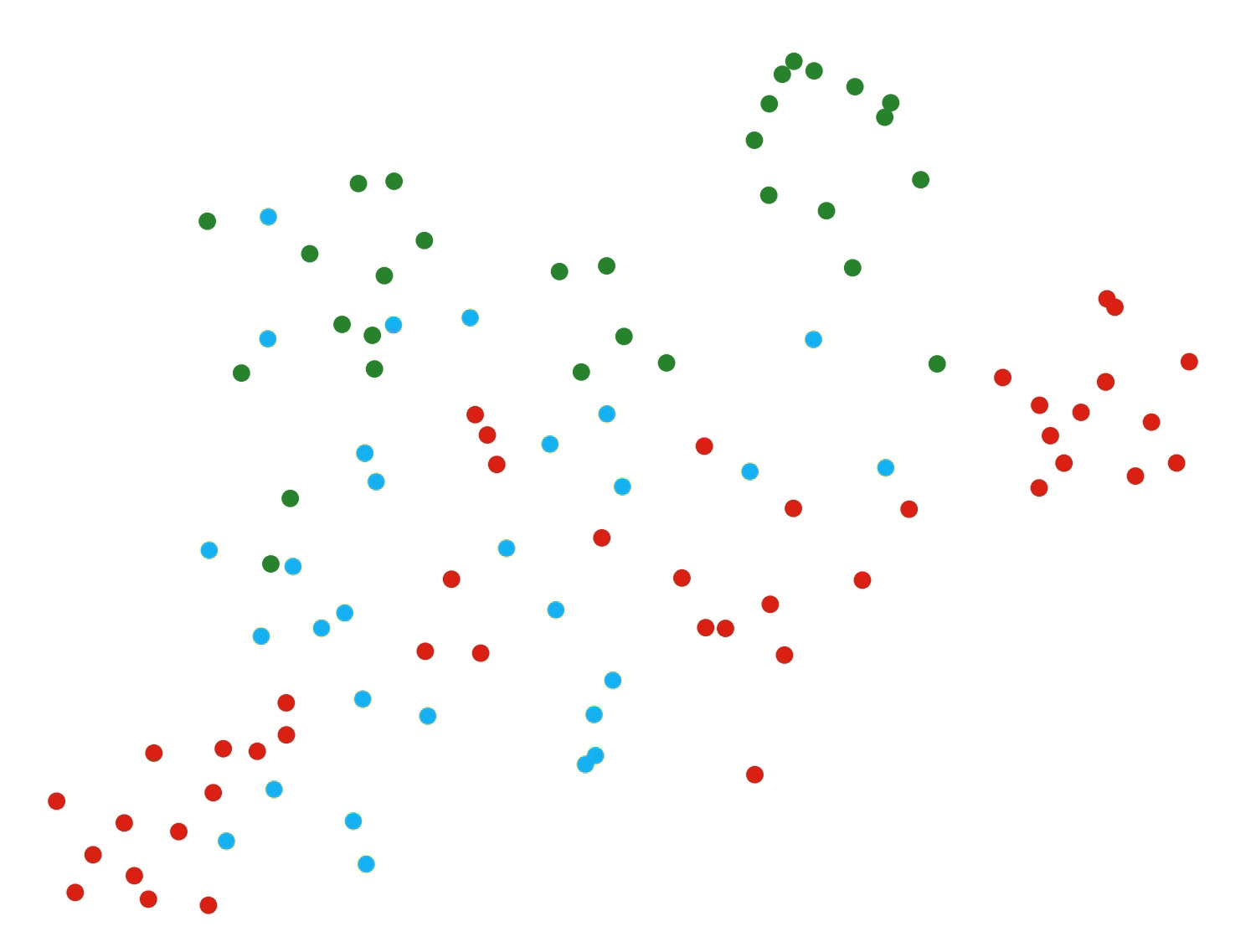}
    }
    \subfigure[ENRC species]{
        \includegraphics[width=0.22\textwidth]{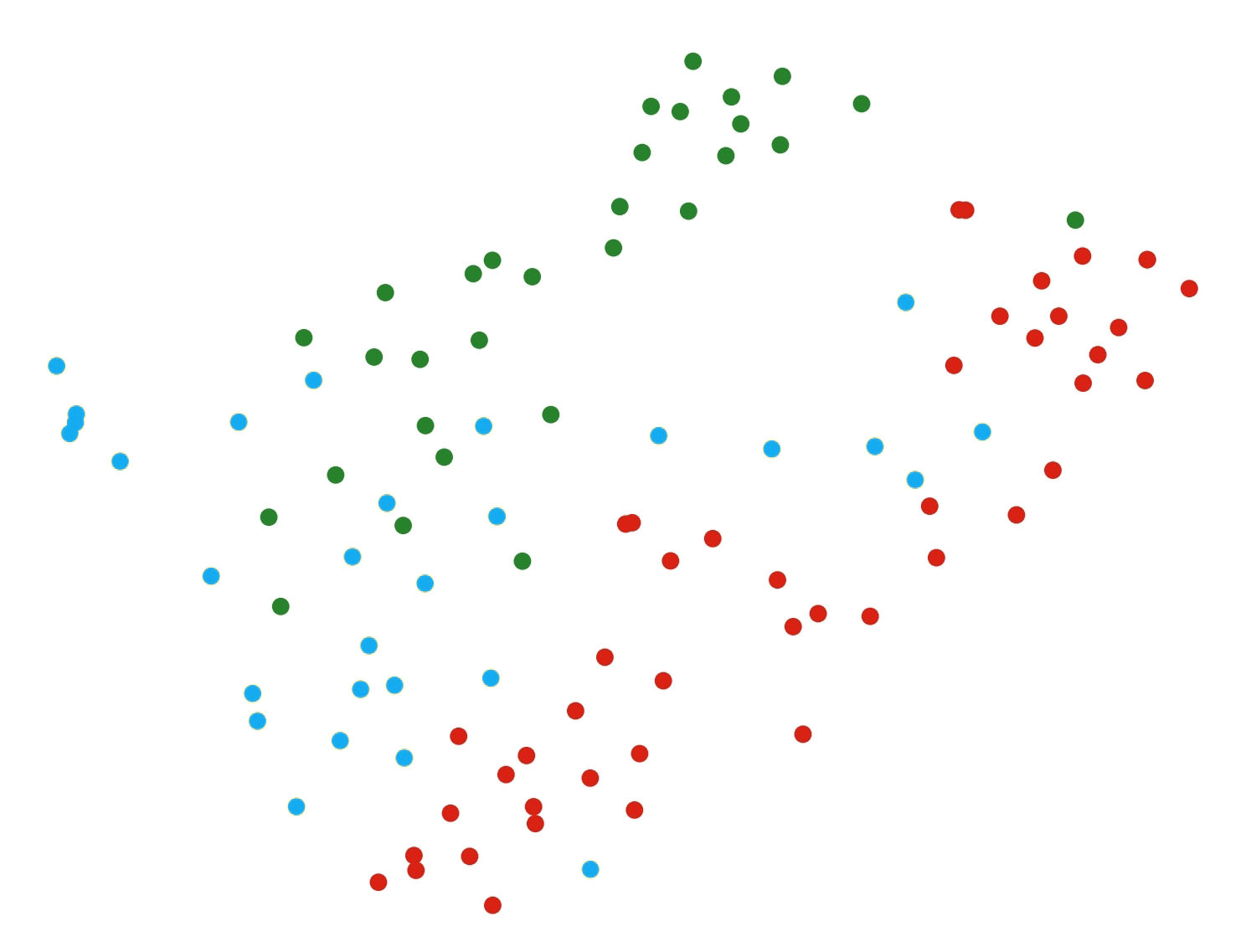}
    }
    \subfigure[AugDMC species]{
        \includegraphics[width=0.22\textwidth]{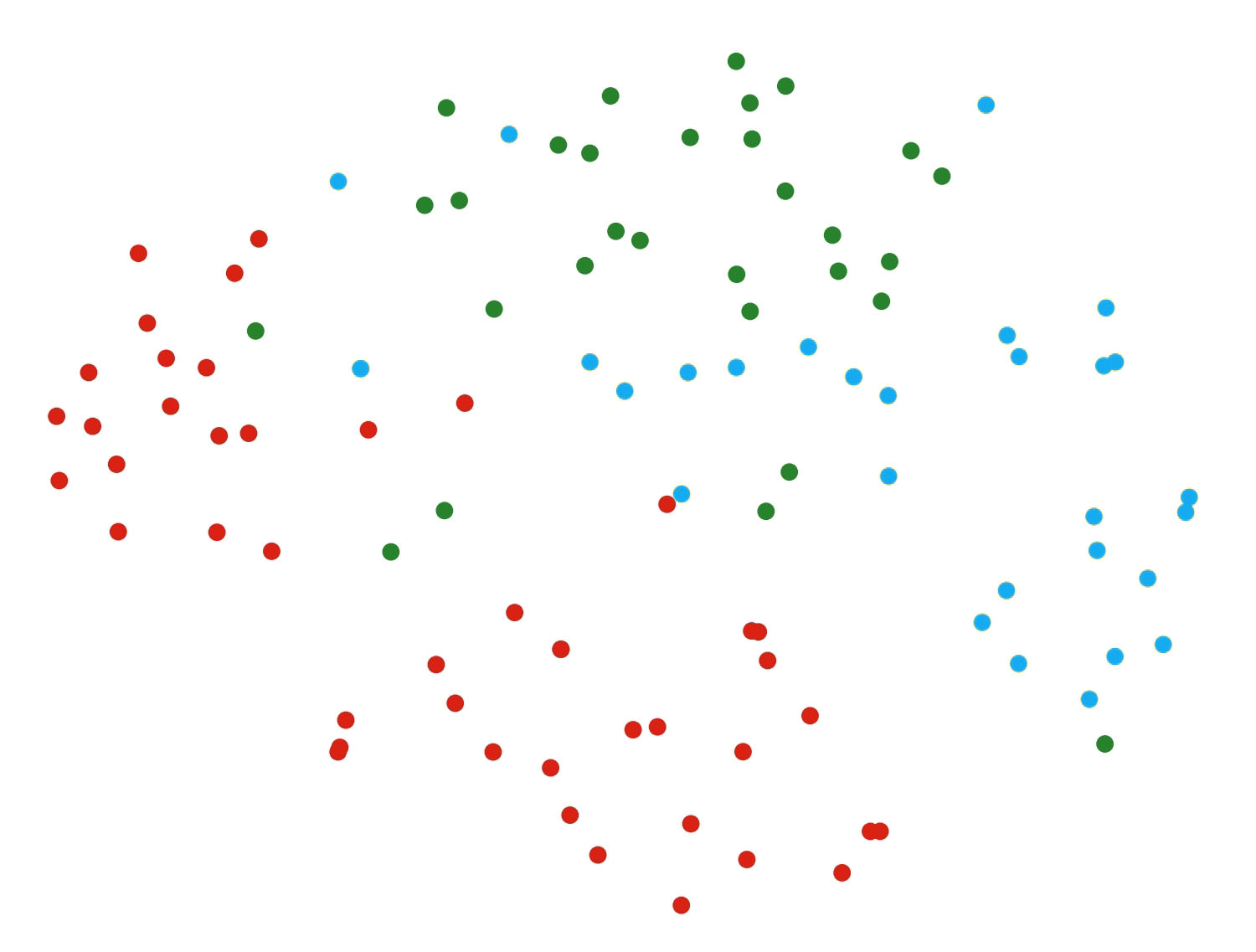}
    }
    \caption{Visualization of image representations on the Fruit dataset. 
    For the results of color clusterings, the red, blue, and green points indicate images with red, yellow, and green labels, respectively.
    For the results of species clusterings, the  red, blue, and green points correspond to images with apple, banana, and grapes labels, respectively.
    }
    \label{fig:vis}
\end{figure*}






\begin{table*}[ht]
    \centering
    \caption{The multiple clusterings performance comparison. The best results are in bold.}
    \resizebox{0.78\textwidth}{!}{
        \begin{tabular}{cc|cc cc cc cc cc}
        \toprule
         \multirow{2}{*}{Dataset}    &   \multirow{2}{*}{Clustering Type}  & \multicolumn{2}{c}{MSC} & \multicolumn{2}{c}{MCV} & \multicolumn{2}{c}{ENRC} & \multicolumn{2}{c}{iMClusts} & \multicolumn{2}{c}{AugDMC} \\
         && NMI & RI & NMI & RI & NMI & RI & NMI & RI & NMI & RI \\
         \midrule
         \multirow{2}{*}{Fruit}   
         & Color   & 0.6886 & 0.8051 & 0.6266 & 0.7685 & 0.7103 & 0.8511 & 0.7351 & 0.8632 & \textbf{0.8517} & \textbf{0.9108}\\
         & Species & 0.1627 & 0.6045 & 0.2733 & 0.6597 & 0.3187 & 0.6536 & 0.3029& 0.6743 & \textbf{0.3546} & \textbf{0.7399}\\
         \midrule
         \multirow{2}{*}{Fruit360}
         & Color   & 0.2544 & 0.6054 & 0.3776 & 0.6791 & 0.4264 & 0.6868 & 0.4097 & 0.6841 & \textbf{0.4594} & \textbf{0.7392} \\
         & Species   & 0.2184 & 0.5805 & 0.2985 & 0.6176 & 0.4142 & 0.6984 & 0.3861& 0.6732 & \textbf{0.5139} & \textbf{0.7430}\\
         \midrule
         \multirow{2}{*}{Card}
         & Order & 0.0807 & 0.7805 & 0.0792 & 0.7128 & 0.1225 & 0.7313 & 0.1144 & 0.7658 & \textbf{0.1440} & \textbf{0.8267} \\
         & Suits & 0.0497 & 0.3587 & 0.0430 & 0.3638 & 0.0676 & 0.3801 & 0.0716 & 0.3715 & \textbf{0.0873} & \textbf{0.4228}\\
         
         \bottomrule
        \end{tabular}
    }
    \label{tab:clustering_res}
\end{table*}
\section{Experiment}
In this section, we conduct experiments to demonstrate our proposal AugDMC. We first introduce our experimental setup, and then present empirical results in comparison to state-of-the-art baseline methods.

\subsection{Experimental Setup}


\paragraph{Datasets}
We conduct the experiments on three image datasets. First, Fruit~\cite{hu2017finding} dataset consists of 105 images
    and has two clusterings, i.e., species and color.
    Specifically,
    it contains three species (i.e., apple, banana, and grape) and three colors (i.e., green, red, and yellow).
    Second, Fruit360\footnote{https://www.kaggle.com/moltean/fruits} dataset contains 4856 images and also has two clusterings, 
    i.e., species (apple, banana, cherry, and grape) and color (red, green, yellow, and maroon). Different from Fruit~\cite{hu2017finding}, images in Fruit360 are with more classes, and thus more complex.
    Third, Card\footnote{https://www.kaggle.com/datasets/gpiosenka/cards-image-datasetclassification} is a dataset of playing card images, which consists of 8,029 images with two clusterings, i.e., rank (Ace, King, Queen, etc.) and suits (clubs, diamonds, hearts, spades).
\begin{figure*}
    \centering
    \subfigure[MSC color]{
        \includegraphics[width=0.22\textwidth]{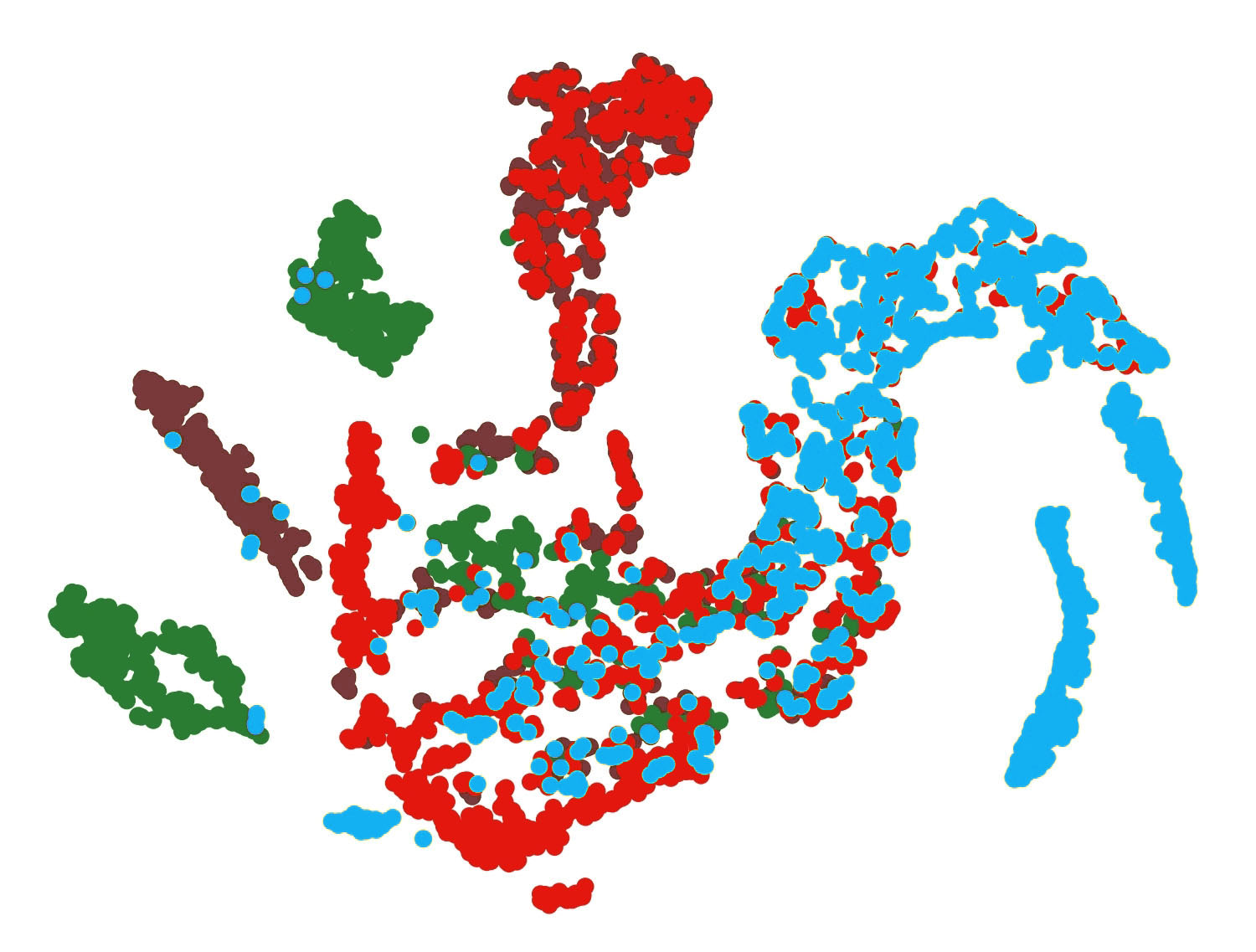}
    }
    \subfigure[MCV color]{
        \includegraphics[width=0.22\textwidth]{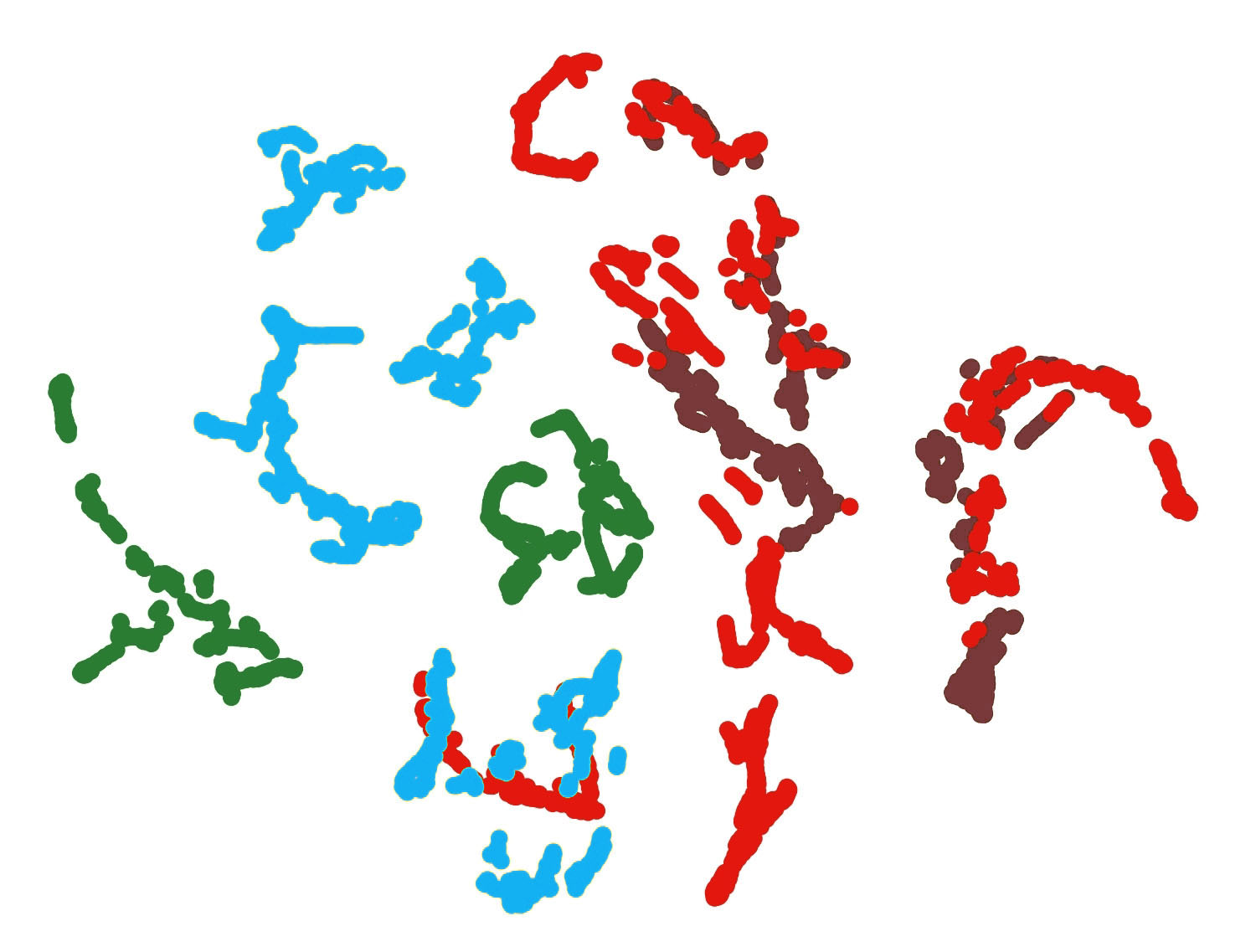}
    }
    \subfigure[ENRC color]{
        \includegraphics[width=0.22\textwidth]{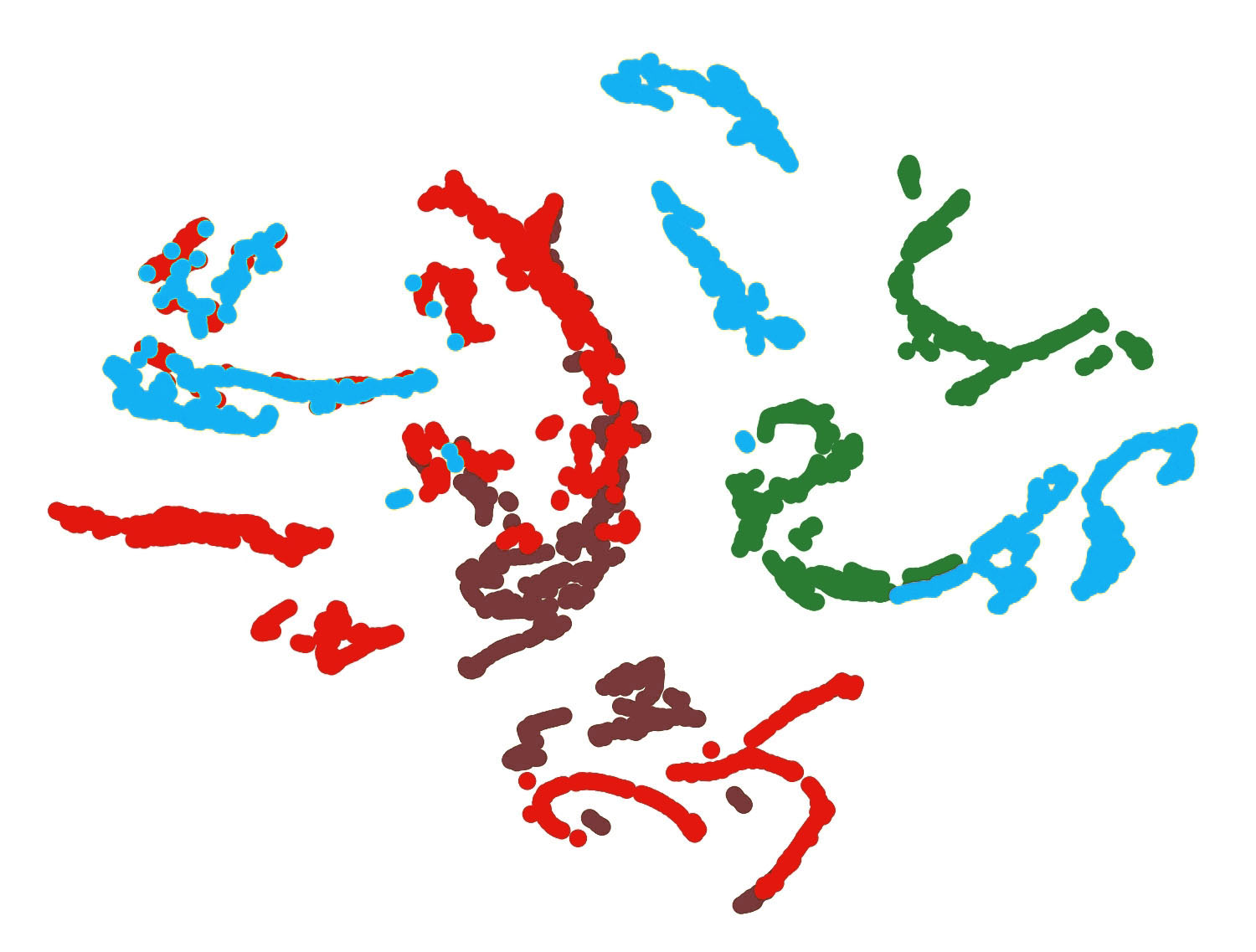}
    }
    \subfigure[AugDMC color]{
        \includegraphics[width=0.22\textwidth]{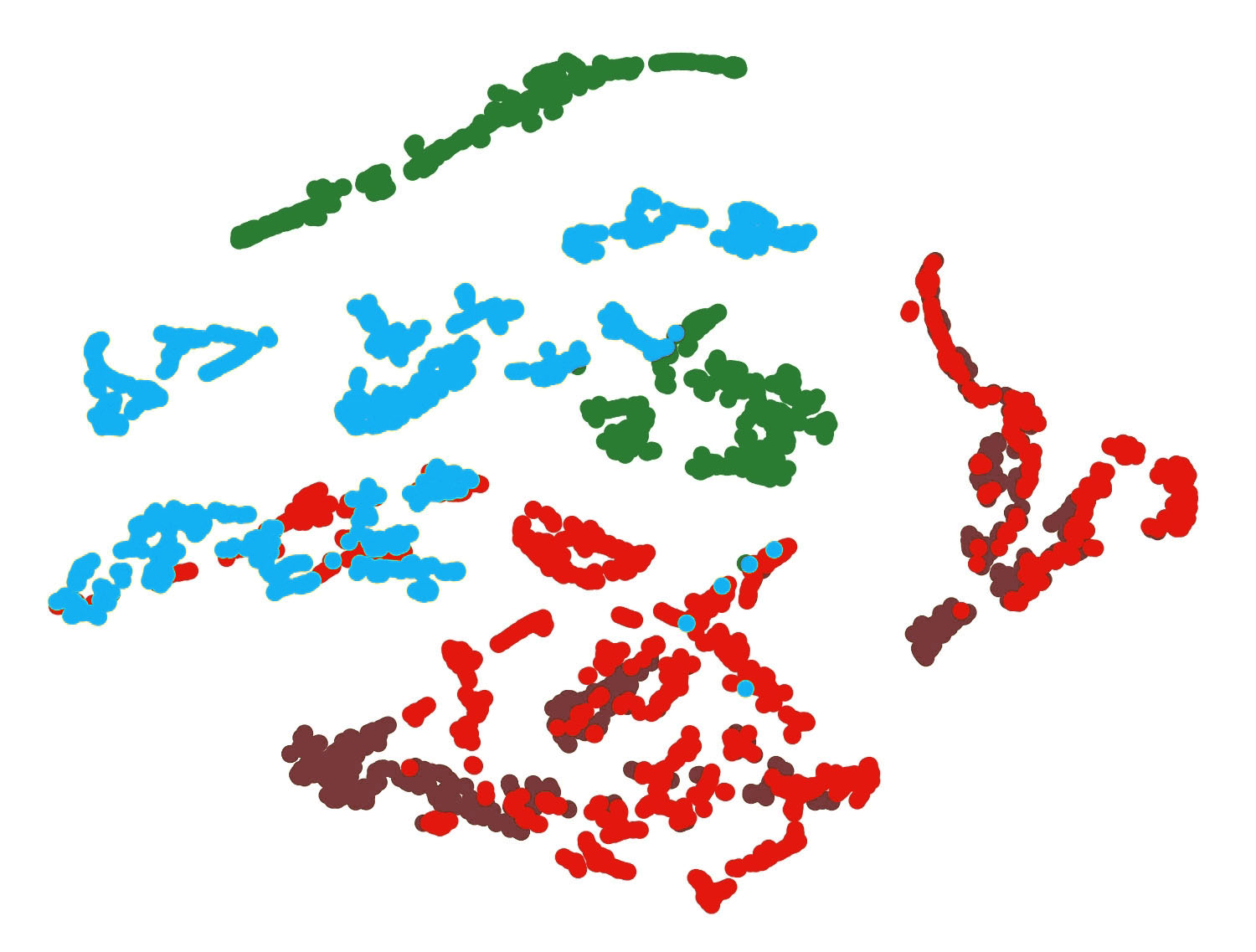}
        
    }\\
    \subfigure[MSC species]{
        \includegraphics[width=0.22\textwidth]{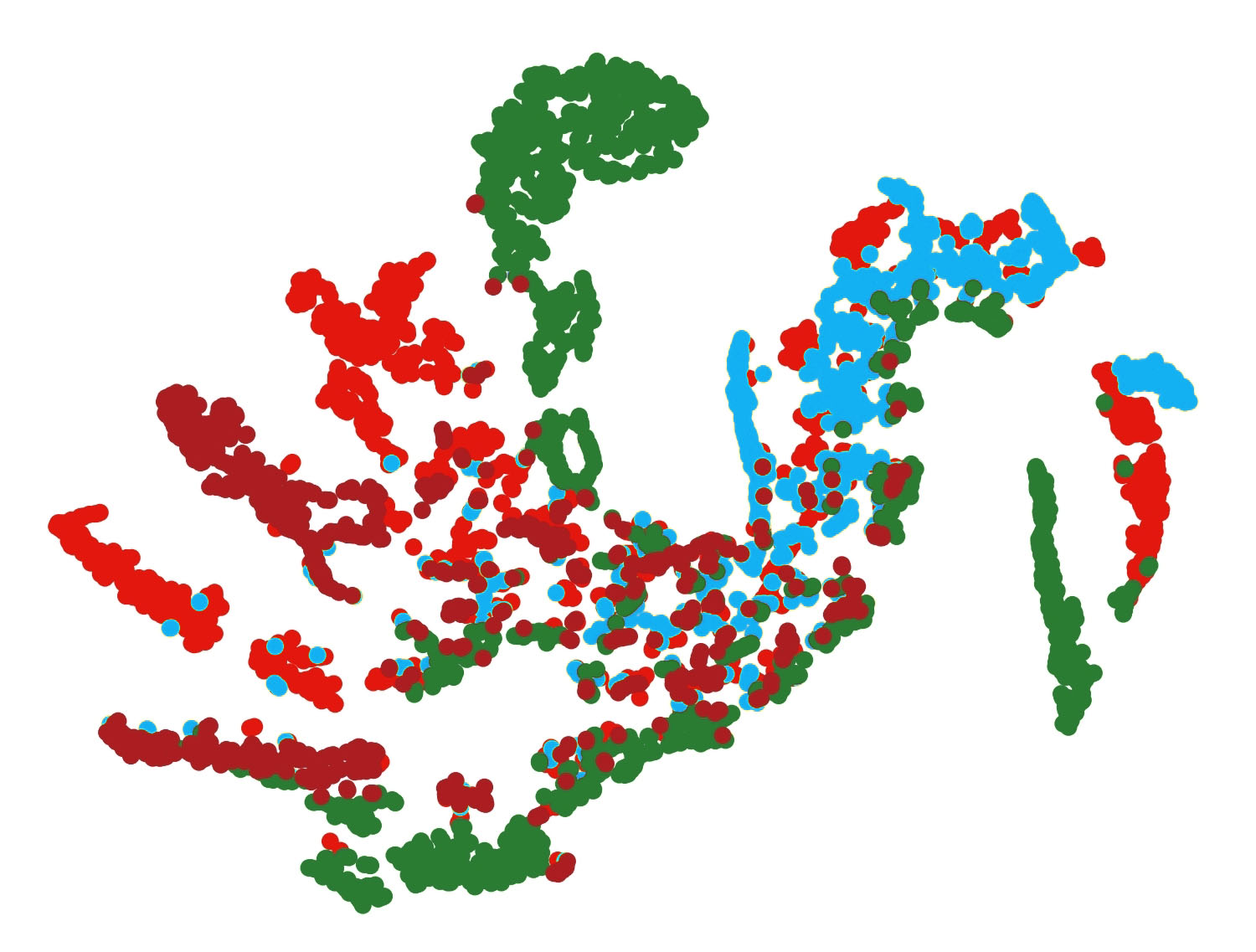}
    }
    \subfigure[MCV species]{
        \includegraphics[width=0.22\textwidth]{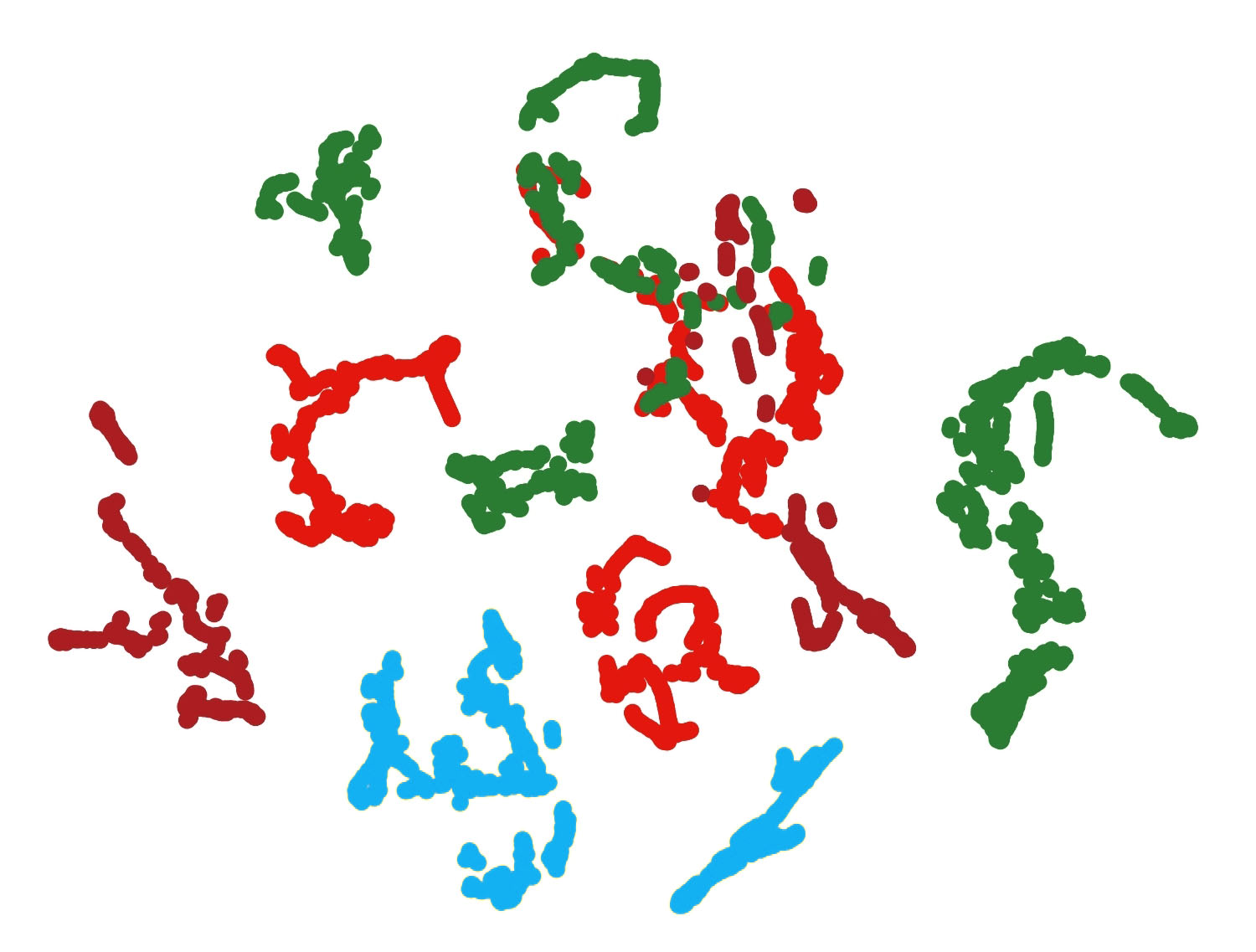}
    }
    \subfigure[ENRC species]{
        \includegraphics[width=0.22\textwidth]{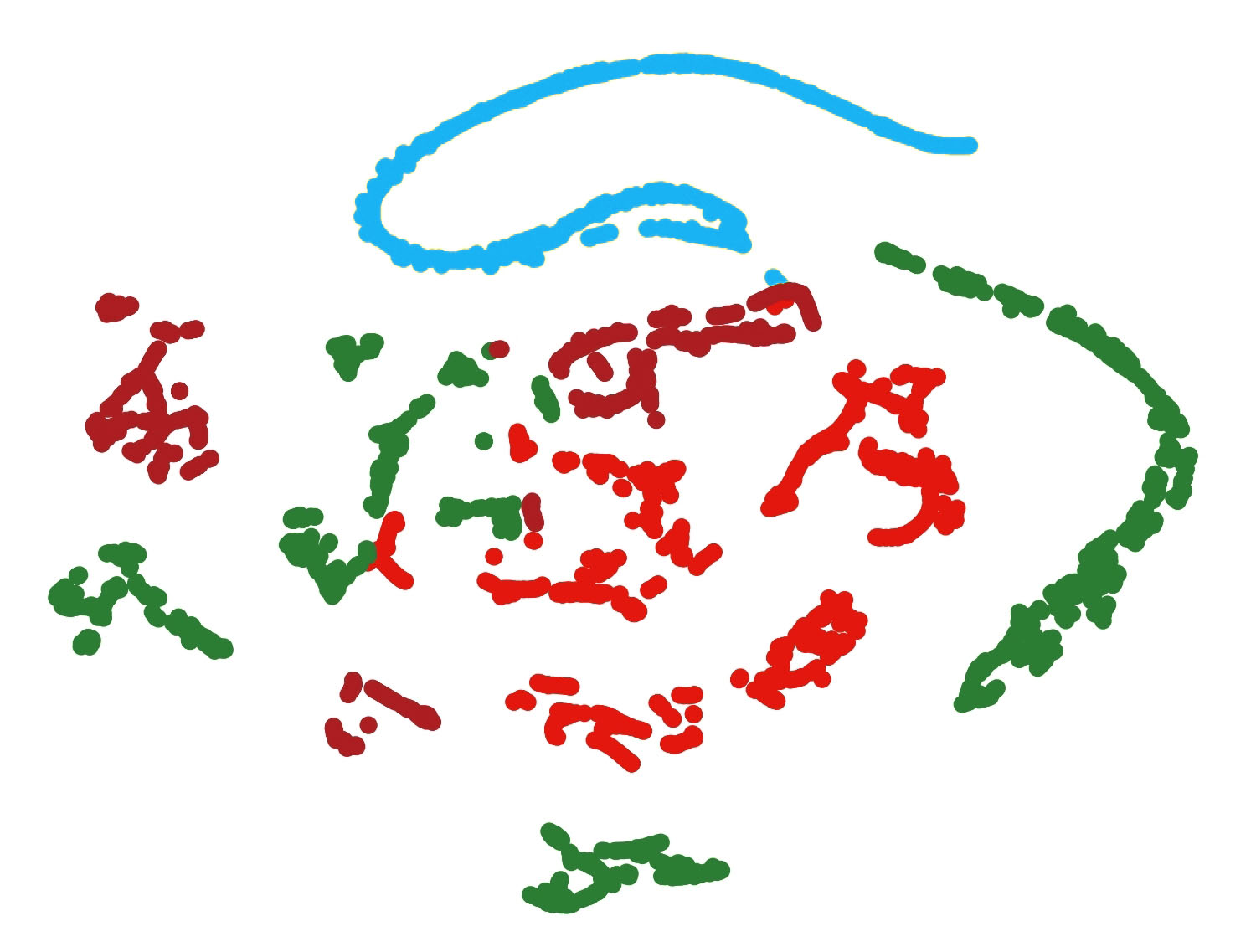}
    }
    \subfigure[AugDMC species]{
        \includegraphics[width=0.22\textwidth]{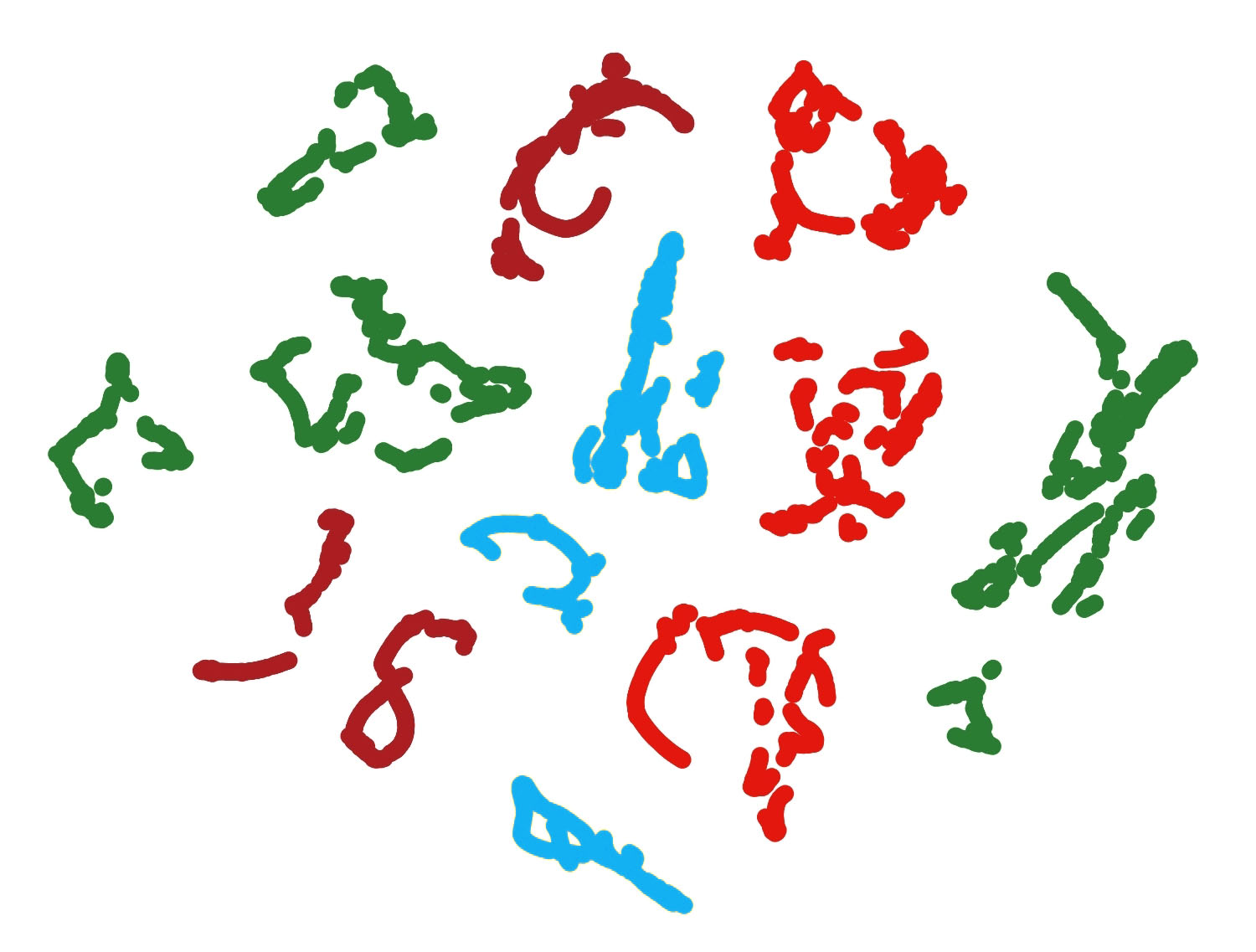}
    }
    \caption{Visualization of image representations on Fruit360 dataset. 
    For the results of color clusterings, 
    the red, blue, green, and maroon points signify images with red, yellow, green, and maroon labels, respectively.
    For the results of species clusterings,
    the images with apple, banana, cherry, and grape labels are marked by red, blue, green, and maroon points, respectively.}
    \label{fig:vis_fruit360}
\end{figure*}

\paragraph{Baselines}
The proposed AugDMC is compared with the following state-of-the-art methods:
    (1) 
    MSC~\cite{hu2017finding} is a traditional multiple clustering method, which considers stability of clusterings and finds multiple clusterings by maximizing the Laplacian eigengap; 
    (2) 
    MCV~\cite{guerin2018improving} considers multiple different
    pre-trained feature extractors as different ``views'' of the same data,
    and designs a multi-input neural network to obtain a better clustering result; 
    (3) 
    ENRC~\cite{miklautz2020deep} is a deep multiple clustering method, which combines auto-encoder and clustering objective function to obtain alternative clusterings; and 
    (4) 
    iMClusts~\cite{ren2022diversified} makes use of the expressive representational power of deep autoencoders and multi-head attention to achieve multiple clusterings.


\paragraph{Implementation Details}
AugDMC uses a common and efficient backbone, that is ResNet-18~\cite{he2016deep}, to do self-supervised representation learning. Several data augmentation methods such as `RandomRotation' and `RandomHorizontalFlip' that will not change the data's perspective are included in all experiments for the effectiveness of representation learning. For the clustering on color, we add data augmentation of `RandomCrop' with a minimum size of half, in which the color perspective should be preserved. For the clustering on species, we add data augmentation of `ColorJitter', in which the species shape should be preserved.

The training process is optimized using Stochastic Gradient Descent (SGD).
All hyperparameters are searched according to the loss of self-supervised representation learning, where
    learning rate is searched in  $\{0.2,0.1,0.05,0.01,0.005,0.0001\}$, 
    weight decay is from$\{0.001,0.0005, 0.0001, 0.00005\}$, and that of
    temperature $\tau$ is from $\{0.8,0.85,0.9,0.95,1.0\}$.
We also set momentum as $0.9$, training epoch as $1000$.
Furthermore,
AugDMC uses early stopping based on the accuracy of the prototype-based classification in the training process.
After that, we can obtain image representations using the input of the last fully connected layer for clustering, which adopts k-means~\cite{lloyd1982least} in the following results. 
We evaluate the clustering performance compared to the ground truth using two commonly used metrics, that is, Normalized Mutual Information (NMI)~\cite{white2004performance} and Rand index (RI)~\cite{rand1971objective}.
We conduct the experiments with a GPU NVIDIA GeForce RTX 2080 Ti.

\subsection{Performance Comparison}

Table~\ref{tab:clustering_res} compares the clustering performance between AugDMC and all other baselines. The best results are in bold.
We can observe that the proposed method achieves the best results in all cases.
These results demonstrate the effectiveness of the proposed method by capturing the concept of interest using a corresponding data augmentation method.
Also,
we can find that the deep multiple clustering models, i.e., MCV, ENRC, iMClusts and AugDMC,
achieve better results than the shallow model, i.e., MSC, in most cases.
This further confirms that deep multiple clustering methods have a more powerful ability in learning image representations to discover multiple clusterings.
Besides,
for the deep multiple clustering models,
AugDMC achieves 7\% to 24\% improvement compared with the baselines,
suggesting the effectiveness of the proposed method.

\subsection{Ablation Study}

\begin{table}
    \centering
    \caption{Performance contribution of each component  in AugDMC.}
    \resizebox{0.5\textwidth}{!}{
        \begin{tabular}{cc|cc cc cc}
        \toprule
         \multirow{2}{*}{Method}    &   {Clustering}  & \multicolumn{2}{c}{Fruit} & \multicolumn{2}{c}{Fruit360}  & \multicolumn{2}{c}{Card} \\
         & Type & NMI & RI & NMI & RI  & NMI & RI  \\
         \midrule
         \multirow{2}{*}{AugDMC} 
         & Color & \textbf{0.8517} & \textbf{0.9108} & \textbf{0.4594} & \textbf{0.7392}  & \textbf{0.1440} & \textbf{0.8267}  \\
         & Species & \textbf{0.3546} & \textbf{0.7399} & \textbf{0.5139} & \textbf{0.7430}  & \textbf{0.0873} & \textbf{0.4228} \\
         \midrule
         \multirow{2}{*}{AugDMC$_{\text{wo}\tau}$}
         & Color   & 0.8472 & 0.8995 & 0.4407 & 0.7177 & 0.1391 & 0.8003 \\
         & Species & 0.3453 & 0.6901 & 0.5042 & 0.7212 & 0.0810 & 0.4029 \\
         \midrule
         \multirow{2}{*}{AugDMC$_\text{woS}$}
         & Color   & 0.8361 & 0.8979 & 0.4387 & 0.7119 & 0.1326 & 0.7892 \\
         & Species & 0.3409 & 0.7017 & 0.4907 & 0.7285 & 0.0726 & 0.3849 \\
         \midrule
         \multirow{2}{*}{AugDMC$_{\text{woS}\tau}$}
         & Color   & 0.8273 & 0.8873 & 0.4302 & 0.7091 & 0.1261 & 0.7624 \\
         & Species & 0.3389 & 0.6817 & 0.4850 & 0.6995 & 0.0687 & 0.3796 \\
         \midrule
         \multirow{2}{*}{AugDMC$_\text{woA}$}
         & Color   & 0.7172 & 0.8549 & 0.4064 & 0.6828 & 0.1057 & 0.7028 \\
         & Species & 0.3084 & 0.6194 & 0.4249 & 0.6806 & 0.0642 & 0.3623 \\
         \midrule
         \multirow{2}{*}{AugDMC$_{\text{woA}\tau}$}
         & Color   & 0.7030 & 0.8456 & 0.3964 & 0.6842 & 0.0993 & 0.7256 \\
         & Species & 0.3035 & 0.5881 & 0.4131 & 0.6775 & 0.0601 & 0.3609 \\
         \bottomrule
        \end{tabular}
    }
    \label{tab:ablation}
\end{table}

To study the contribution of each component in AugDMC (i.e., the temperature parameter $\tau$ in prototype-based representation learning, data augmentation, and stable optimization strategy), we conduct an ablation study in this subsection. 
Specifically, we remove the above components from AugDMC and obtain three variants named AugDMC$_{\text{wo}\tau}$, AugDMC$_\text{woA}$, and AugDMC$_\text{woS}$, respectively.
Note that we set $\tau=1$ for AugDMC$_{\text{wo}\tau}$.
Besides, 
we further remove the combination of the temperature in prototype-based representation learning and stable optimization strategy, as well as the  combination of the temperature in prototype-based representation learning and data augmentation,
namely AugDMC$_{\text{woS}\tau}$ and AugDMC$_{\text{woA}\tau}$, respectively.

The results are shown in Table~\ref{tab:ablation}.
We can find that AugDMC always achieves the best performance,
indicating the effectiveness of prototype-based representation learning, data augmentation, and stable optimization strategy.
Also,
AugDMC$_{\text{woS}\tau}$ and AugDMC$_{\text{woA}\tau}$ perform worse than AugDMC$_{\text{wo}\tau}$, AugDMC$_\text{woS}$, and AugDMC$_\text{woA}$.
This indicates that the combination of these components is useful for the proposed method. 

\begin{figure}[ht]
\centering
\subfigure[{\scriptsize Clustering by color on Fruit}]{
            \begin{tikzpicture}[font=\Large, scale=0.45]
                \begin{axis}[
                    legend cell align={left},
                    legend style={nodes={scale=1.0, transform shape}, at={(0.03,0.5)},anchor=west},
                    xlabel={temperature $\tau$},
                    xtick pos=left,
                    tick label style={font=\large},
                    ylabel style={font=\large},
                    xtick={0.8, 0.85, 0.9, 0.95, 1.0},
                    xticklabels={$0.8$,$0.85$,$0.9$,$0.95$,$1.0$},
                    ytick={0.78, 0.82,0.87,0.92},
                    yticklabels={ $0.78$,$0.82$,$0.87$,$0.92$},
                    legend pos= south east,
                    ymajorgrids=true,
                    grid style=dashed
                ]
                \addplot[
                    color=purple,
                    dotted,
                    mark options={solid},
                    mark=diamond*,
                    line width=1.5pt,
                    mark size=2pt
                    ]
                    coordinates {
                    (0.8, 0.7853)
                    (0.85, 0.8319)
                    (0.9, 0.8517)
                    (0.95, 0.8497)
                    (1.0, 0.8472)
                    };
                    \addlegendentry{NMI}
                \addplot[
                    color=blue,
                    dotted,
                    mark options={solid},
                    mark=*,
                    line width=1.5pt,
                    mark size=2pt
                    ]
                    coordinates {
                    (0.8, 0.8535)
                    (0.85, 0.8697)
                    (0.9, 0.9108)
                    (0.95, 0.9087)
                    (1.0, 0.8995)
                    };
                    \addlegendentry{RI}
                \end{axis}
                \end{tikzpicture}
    }
\subfigure[{\scriptsize Clustering by species on Fruit}]{
            \begin{tikzpicture}[font=\Large, scale=0.45]
                \begin{axis}[
                    legend cell align={left},
                    legend style={nodes={scale=1.0, transform shape}, at={(0.03,0.5)},anchor=west},
                    xlabel={temperature $\tau$},
                    xtick pos=left,
                    tick label style={font=\large},
                    ylabel style={font=\large},
                    xtick={0.8, 0.85, 0.9, 0.95, 1.0},
                    xticklabels={$0.8$,$0.85$,$0.9$,$0.95$,$1.0$},
                    ytick={0.32, 0.42,0.53,0.64,0.74},
                    yticklabels={$0.32$, $0.42$,$0.53$,$0.64$,$0.74$},
                    ymajorgrids=true,
                    grid style=dashed
                ]
                \addplot[
                    color=purple,
                    dotted,
                    mark options={solid},
                    mark=diamond*,
                    line width=1.5pt,
                    mark size=2pt
                    ]
                    coordinates {
                    (0.8, 0.3282)
                    (0.85, 0.3295)
                    (0.9, 0.3546)
                    (0.95, 0.3519)
                    (1.0, 0.3453)
                    };
                    \addlegendentry{NMI}
                \addplot[
                    color=blue,
                    dotted,
                    mark options={solid},
                    mark=*,
                    line width=1.5pt,
                    mark size=2pt
                    ]
                    coordinates {
                    (0.8, 0.6835)
                    (0.85, 0.7057)
                    (0.9, 0.7399)
                    (0.95, 0.7237)
                    (1.0, 0.6901)
                    };
                    \addlegendentry{RI}
                \end{axis}
                \end{tikzpicture}
    } 
\subfigure[{\scriptsize Clustering by color on Fruit360}]{
            \begin{tikzpicture}[font=\Large, scale=0.45]
                \begin{axis}[
                    legend cell align={left},
                    legend style={nodes={scale=1.0, transform shape}, at={(0.03,0.5)},anchor=west},
                    xlabel={temperature $\tau$},
                    xtick pos=left,
                    tick label style={font=\large},
                    ylabel style={font=\large},
                    xtick={0.8, 0.85, 0.9, 0.95, 1.0},
                    xticklabels={$0.8$,$0.85$,$0.9$,$0.95$,$1.0$},
                    ytick={0.43, 0.50, 0.59,0.68,0.75},
                    yticklabels={ $0.43$,$0.50$,$0.59$,$0.68$,$0.75$},
                    ymajorgrids=true,
                    grid style=dashed
                ]
                \addplot[
                    color=purple,
                    dotted,
                    mark options={solid},
                    mark=diamond*,
                    line width=1.5pt,
                    mark size=2pt
                    ]
                    coordinates {
                    (0.8, 0.4383)
                    (0.85, 0.4431)
                    (0.9, 0.4579)
                    (0.95, 0.4594)
                    (1.0, 0.4407)
                    };
                    \addlegendentry{NMI}
                \addplot[
                    color=blue,
                    dotted,
                    mark options={solid},
                    mark=*,
                    line width=1.5pt,
                    mark size=2pt
                    ]
                    coordinates {
                    (0.8, 0.7234)
                    (0.85, 0.7206)
                    (0.9, 0.7352)
                    (0.95, 0.7392)
                    (1.0, 0.7177)
                    };
                    \addlegendentry{RI}
                \end{axis}
                \end{tikzpicture}
    }
\subfigure[{\scriptsize Clustering by species on Fruit360}]{
            \begin{tikzpicture}[font=\Large, scale=0.45]
                \begin{axis}[
                    legend cell align={left},
                    legend style={nodes={scale=1.0, transform shape}, at={(0.03,0.5)},anchor=west},
                    xlabel={temperature $\tau$},
                    xtick pos=left,
                    tick label style={font=\large},
                    ylabel style={font=\large},
                    xtick={0.8, 0.85, 0.9, 0.95, 1.0},
                    xticklabels={$0.8$,$0.85$,$0.9$,$0.95$,$1.0$},
                    ytick={0.32, 0.43,0.53,0.63,0.75},
                    yticklabels={$0.32$, $0.43$,$0.53$,$0.63$,$0.75$},
                    ymajorgrids=true,
                    grid style=dashed
                ]
                \addplot[
                    color=purple,
                    dotted,
                    mark options={solid},
                    mark=diamond*,
                    line width=1.5pt,
                    mark size=2pt
                    ]
                    coordinates {
                    (0.8, 0.3282)
                    (0.85, 0.3295)
                    (0.9, 0.3519)
                    (0.95, 0.3546)
                    (1.0, 0.3453)
                    };
                    \addlegendentry{NMI}
                \addplot[
                    color=blue,
                    dotted,
                    mark options={solid},
                    mark=*,
                    line width=1.5pt,
                    mark size=2pt
                    ]
                    coordinates {
                    (0.8, 0.7183)
                    (0.85, 0.7257)
                    (0.9, 0.7377)
                    (0.95, 0.7430)
                    (1.0, 0.7212)
                    };
                    \addlegendentry{RI}
                \end{axis}
                \end{tikzpicture}
    } 
    \caption{Results of parameter sensitivity.}
    \label{fig:parameter}
\end{figure}
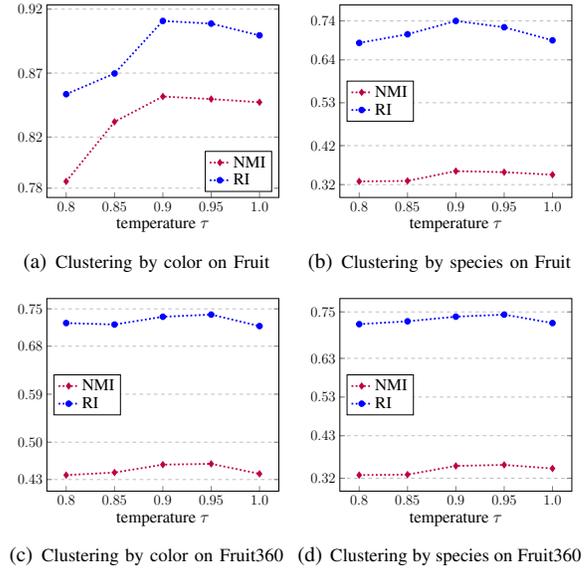

\paragraph{Parameter Sensitivity}

Moreover, we investigate the effect of the temperature $\tau$ in AugDMC on Fruit and Fruit360 datasets.
The results of AugDMC under varying $\tau$ are shown in Fig.~\ref{fig:parameter}.
With the increase of the value $\tau$,
the performance of AugDMC first improves and then drops on both datasets. This suggests that a better choice of the temperature can further improve the performance, however the change is not big in most cases.

\subsection{Visualization}

To further demonstrate the effectiveness of the proposed method, in this subsection, we visualize the learned representations on the Fruit and Fruit360 datasets using t-SNE~\cite{van2008visualizing} to compare different methods.
The results on the Fruit and Fruit360 dataset are shown in Figs.~\ref{fig:vis} and \ref{fig:vis_fruit360}, respectively.
Comparing these results,
we can find that the representations with different labels learned by MSC, MCV, and ENRC are mixed with each other, 
while AugDMC distinguishes all the categories with a more clear boundary.




\section{Conclusion}
In this paper, we study the problem of deep multiple clusterings for images and propose a novel augmentation guided method, named AugDMC, to flexibly and efficiently capture users' interest.
Specifically,
we perturb given images using augmentations to control the aspect to be clustered, where the corresponding image representations can be obtained through prototype-based representation learning with a stable optimization strategy.
Experiments on three real-world datasets demonstrate the effectiveness of the proposed method. Our representation learning is independent from any clustering constraints, which makes the learned representations of different aspects flexible for other downstream tasks as a future direction. 

\section{Acknowledgement}

Yao and Hu's research is supported in part by Advata Gift Funding. All opinions, findings, conclusions and recommendations in this paper are those of the author and do not necessarily reflect the views of the funding agencies.

\bibliographystyle{IEEEtran}
\bibliography{ref}

\end{document}